\documentclass[runningheads]{llncs}
\usepackage{graphicx}
\usepackage{comment}
\usepackage{amsmath,amssymb} 
\usepackage{color}

\usepackage{epsfig}
\usepackage{bbm}
\usepackage{algorithm}
\usepackage{enumitem}
\usepackage{algorithmic}
\usepackage{xcolor}
\usepackage{pifont}
\usepackage{hyperref}
\usepackage{subfigure}
\newcommand{\E}{\mathbb{E}}
\newcommand{\etal}{\textit{et al}.}
\newcommand{\ie}{\textit{i}.\textit{e}.}
\newcommand{\eg}{\textit{e}.\textit{g}.}

\begin{document}
\pagestyle{headings}
\mainmatter
\def\ECCVSubNumber{3029}  

\title{A Competence-aware Curriculum for Visual Concepts Learning via Question Answering} 

\titlerunning{Competence-aware Curriculum for Visual Concepts Learning}
%
\author{Qing Li\orcidID{0000-0003-1185-5365} \and
Siyuan Huang\orcidID{0000-0003-1524-7148} \and
Yining Hong\orcidID{0000-0002-0518-2099} \and
Song-Chun Zhu\orcidID{0000-0002-1925-5973}}
\authorrunning{Q. Li et al.}
%
\institute{UCLA Center for Vision, Cognition, Learning, and Autonomy (VCLA)
\\ \email{\{liqing, huangsiyuan, yininghong\}@ucla.edu}, \email{sczhu@stat.ucla.edu}
}

\maketitle

\begin{abstract}
Humans can progressively learn visual concepts from easy to hard questions. To mimic this efficient learning ability, we propose a competence-aware curriculum for visual concept learning in a question-answering manner. Specifically, we design a neural-symbolic concept learner for learning the visual concepts and a multi-dimensional Item Response Theory (mIRT) model for guiding the learning process with an adaptive curriculum. The mIRT effectively estimates the concept difficulty and the model competence at each learning step from accumulated model responses. The estimated concept difficulty and model competence are further utilized to select the most profitable training samples. Experimental results on CLEVR show that with a competence-aware curriculum, the proposed method achieves state-of-the-art performances with superior data efficiency and convergence speed. Specifically, the proposed model only uses \textbf{40\% of training data} and converges \textbf{three times faster} compared with other state-of-the-art methods. 

\end{abstract}

\section{Introduction}

\begin{figure}[t]
    \centering {\includegraphics[width=1\textwidth]{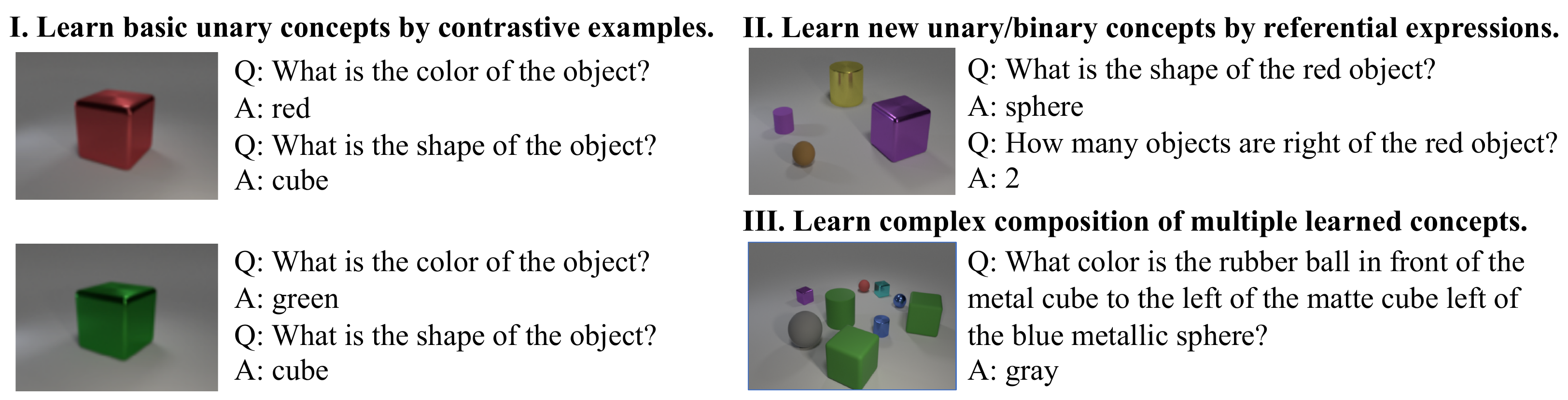}}
    \caption{The incremental learning of visual concepts in a question-answering manner. Three difficulty levels can be categorized into I) unary concepts from simple questions, II) binary (relational) concepts based on the learned concepts, and III) compositions of visual concepts from comprehensive questions.}
    \label{fig:intro}
\end{figure}

Humans excel at learning visual concepts and their compositions in a question-answering manner~\cite{fazly2010probabilistic,chrupala2015learning,gauthier2018word}, which requires a joint understanding of vision and language. The essence of such learning skill is the superior capability to connect linguistic symbols (words/phrases) in question-answer pairs with visual cues (appearance/geometry) in images. 
Imagine a person without prior knowledge of colors is presented with two contrastive examples in Figure~\ref{fig:intro}-I. The left images are the same except for color, and the right question-answer pairs differ only in the descriptions about color. By assuming that the differences in the question-answer pairs capture the differences in appearances, he can learn the concept of color and the appearance of specific colors (\ie, red and green). Besides learning the basic unary concepts from contrastive examples, compositional relations from complex questions consisting of multiple concepts can be further learned, as shown in Figure~\ref{fig:intro}-II and -III.

Another crucial characteristic of the human learning process is to start \textit{small} and learn \textit{incrementally}. More specifically, the human learning process is well-organized with a curriculum that introduces concepts progressively and facilitates the learning of new abstract knowledge by exploiting learned concepts. A good curriculum serves as an experienced teacher. By ranking and selecting examples according to the learning state, it can guide the training process of the learner (student) and significantly increase the learning speed. This idea is originally examined in animal training as \textit{shaping}~\cite{skinner1958reinforcement,peterson2004day,krueger2009flexible} and then applied to machine learning as \textit{curriculum learning}~\cite{elman1993learning,bengio2009curriculum,Graves2017AutomatedCL,Guo2018CurriculumNetWS,Pentina2014CurriculumLO}.

Inspired by the efficient curriculum, Mao \etal ~\cite{mao2019neuro} proposes a neural-symbolic approach to learn visual concepts with a \textit{fixed} curriculum. Their approach learns from image-question-answer triplets and does not require annotation on images or programs generated from questions. The model is trained with a manually-designed curriculum that includes four stages: (1) learning unary visual concepts; (2) learning relational concepts; (3) learning more complex questions with visual perception fixed; (4) joint fine-tuning all modules. They select questions for each stage by the depths of the latent programs. Their curriculum heavily relies on the manually-designed heuristic that measures the question difficulty and discretizes the curriculum. Such heuristic suffers from three limitations. First, it ignores the variance of difficulties for questions with the same program depths, where different concepts might have various difficulties. Second, the manually-designed curriculum relies on strong human prior knowledge for the difficulties, while such prior may conflict with the inherent difficulty distribution of the training examples.  Last but most importantly, it neglects the progress of the learner that evolves along with the training process. More specifically, the order of training samples in the curriculum is nonadjustable based on the model state. This scheme is in stark contrast to the way that humans learn -- by \textit{actively} selecting learning samples based on our current learning state, instead of \textit{passively} accepting specific training samples. A desirable learning system should be capable of automatically adjusting the curriculum during the learning process without requiring any prior knowledge, which makes the learning procedure more efficient with less data redundancy and faster convergence speed.

To address these issues and mimic human ability in adaptive learning, we propose a \textbf{competence-aware} curriculum for visual concept learning via question answering, where competence represents the capability of the model to recognize each concept. The proposed approach utilizes multi-dimensional Item Response Theory (mIRT) to estimate the \textbf{concept difficulty} and \textbf{model competence} at each learning step from accumulated model responses. Item Response Theory (IRT)~\cite{baker2001basics,baker2004item} is a widely adopted method in psychometrics that estimates the human ability and the item difficulty from human responses on various items. We extend the IRT to a mIRT that matches the compositional nature of visual reasoning, and apply variational inference to get a Bayesian estimation for the parameters in mIRT. Based on the estimations of concept difficulty and model competence, we further define a continuous adaptive curriculum (instead of a discretized fixed regime) that selects the most profitable training samples according to the current learning state. More specifically, the learner can filter out samples with either too naive or too challenging questions. These questions bring either negligible or sharp gradients to the learner, which makes it slower and harder to converge. 

With the proposed competence-aware curriculum, the learner can address the aforementioned limitations brought by a fixed curriculum with the following advantages: 

\begin{enumerate}[leftmargin=*,noitemsep,nolistsep]
    \item The concept difficulty and the model competence at each learning step can be inferred effectively from accumulated model responses. It enables the model to distinguish difficulties among various concepts and be aware of its own capability for recognizing these concepts. 
    \item The question difficulty can be calculated with the estimated concept difficulty and model competence without requiring any heuristics.
    \item The adaptive curriculum significantly contributes to the improvement of learning efficiency by relieving the data redundancy and accelerating the convergence, as well as the improvement of the final performance.
\end{enumerate}

We explore the proposed method on the CLEVR dataset~\cite{johnson2017clevr}, an artificial universe where visual concepts are clearly defined and less correlated. We opt for this synthetic environment because there is little prior work on curriculum learning for visual concepts and there lacks a clear definition of visual concepts in real-world setting.
CLEVR allows us to perform controllable diagnoses of the proposed mIRT model in building an adaptive curriculum. \autoref{sec_discussion} further discusses the potentials and challenges of generalizing our method to other domains such as real-world images and natural language processing.

Experimental results show that the visual concept learner with the proposed competence-aware curriculum converges three times faster and consumes only $40\%$ of the training data while achieving similar or even higher accuracy compared with other state-of-the-art models. We also evaluate individual modules in the proposed method and demonstrate their efficacy in \autoref{sec:exp}.

\section{Related Work}
\subsection{Neural-symbolic Visual Question Answering}
Visual question answering (VQA)~\cite{Malinowski:2014:MAQ:2968826.2969014,tu2014joint,qi2015restricted,johnson2017clevr,gan2017vqs} is a popular task for gauging the capability of visual reasoning systems.
Some recent studies~\cite{Andreas2015NeuralMN,andreas-etal-2016-learning,Hu2017LearningTR,InferringFeifei2017,yi2020clevrer} focus on learning the neural module networks (NMNs) on the CLEVR dataset. 
NMNs translate questions into programs, which are further executed over image features to predict answers. The program generator is typically trained on human annotations. Several recent works target on reducing the supervision or increasing the generalization ability to new tasks in NMNs. For example, Johnson \etal~\cite{InferringFeifei2017} replaces the hand-designed syntactic parsers by a learned program generator. Neural-Symbolic VQA~\cite{yi2018neural} explores an object-based visual representation and uses a symbolic executor for inferring the answer. 
Neural-symbolic concept learner~\cite{mao2019neuro} uses a symbolic reasoning process and manually-defined curriculum to bridge the learning of visual concepts, words, and the parsing of questions without explicit annotations. In this paper, we build our model on the neural-symbolic concept learner~\cite{mao2019neuro} and learn an adaptive curriculum to select the most profitable training samples. 

Learning-by-asking (LBA)~\cite{Misra2017LearningBA} proposes an interactive learning framework that allows the model to actively query an oracle and discover an easy-to-hard curriculum. LBA uses the expected accuracy improvement over candidate answers as an informativeness measure to pick questions. However, it is costly to compute the expected accuracy improvement for sampled questions since it requires to process all the questions and images through a VQA model. Moreover, the expected accuracy improvement cannot help to learn which specific component of the question contributes to the performance, especially while learning from the answers with little information such as ``yes/no''. In contrast, we select questions by explicitly modeling the difficulty of visual concepts, combined with model competence to infer the difficulty of each question.

\subsection{Curriculum Learning and Machine Teaching}
The competence-aware curriculum in our work is related to \textit{curriculum learning}~\cite{bengio2009curriculum,spitkovsky2010baby,tsvetkov2016learning,Graves2017AutomatedCL,sachan2016easy,Pentina2014CurriculumLO,Guo2018CurriculumNetWS,Platanios2019CompetencebasedCL} and \textit{machine teaching}~\cite{zhu2015machine,zhu2018overview,liu2017iterative,Dasgupta2019TeachingAB,Mansouri2019PreferenceBasedBA,fan2018learning,wu2018learning}. \textit{Curriculum learning} is firstly proposed by Bengio \etal~\cite{bengio2009curriculum} and demonstrates that a dataset order from easy instances to hard ones benefits learning process. The measures of hardness in curriculum learning approaches are usually determined by hand-designed heuristics~\cite{spitkovsky2010baby,tsvetkov2016learning,sachan2016easy,mao2019neuro}. Graves \etal~\cite{Graves2017AutomatedCL} explore learning signals based on the increase rates in prediction accuracy and network complexity to adjust data distributions along with training. Self-paced learning~\cite{kumar2010self,jiang2014self,jiang2015self,sachan2016easy} quantifies the sample hardness by the training loss and formulates curriculum learning as an optimization problem by jointly modeling the sample selection and the learning objective. These hand-designed heuristics are usually task-specific without any generalization ability to other domains.

\textit{Machine teaching}~\cite{zhu2015machine,zhu2018overview,liu2017iterative}
introduces a teacher model that receives feedback from the student model and guides the learning of the student model accordingly. Zhu \etal~\cite{zhu2015machine,zhu2018overview} assume that the teacher knows the ground-truth model (\ie, the Oracle) beforehand and constructs a minimal training set for the student model. The recent works \textit{learning to teach}~\cite{fan2018learning,wu2018learning} break this strong assumption of the existence of the oracle model and endow the teacher with the capability of learning to teach via a reinforcement learning framework. 

Our work explores curriculum learning in visual reasoning,  which is highly compositional and more complex than tasks studied before. Different from previous works, our method requires neither hand-designed heuristics nor an extra teacher model. We combine the idea of \textit{competence} with curriculum learning and propose a novel mIRT model that estimates the concept difficulty and model competence from accumulated model responses. 
\section{Methodology}

\begin{figure}[!tb]
    \centering {\includegraphics[width=1\textwidth]{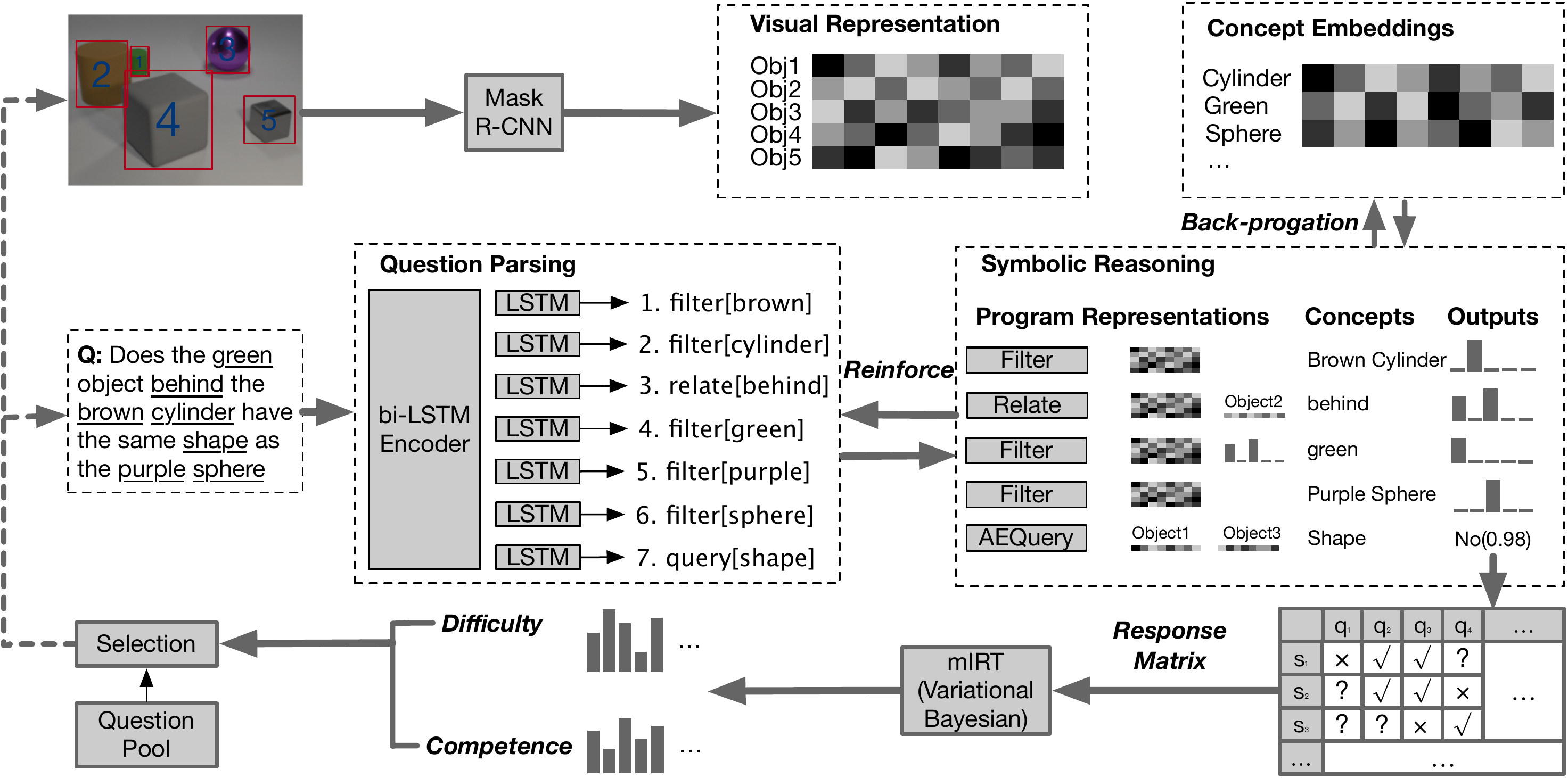}}
    \caption{The overview of the proposed approach. We use neural symbolic reasoning as a bridge to jointly learn concept embeddings and question parsing. The model responses in the training process are accumulated to estimate concept difficulty and model competence at each learning step with mIRT. The estimations help to select appropriate training samples for the current model. In the response matrix,`\checkmark' or `\ding{53}' denotes that the snapshot predicts a correct or wrong answer, and `?' means the snapshot has no response to this question. }
    \label{fig:framework}
\end{figure}

In this section, we will discuss the proposed competence-aware curriculum for visual concept learning, as also shown in Figure~\ref{fig:framework}. We first describe a neural-symbolic approach to learn visual concepts from image-question-answer triplets. Next, we introduce the background of IRT model and discuss how we derive a mIRT model for estimating concept difficulty and model competence. Finally, we present how to select training samples based on the estimated concept difficulty and model competence to make the training process more efficient.

\subsection{Neural-Symbolic Concept Learner}
We briefly describe the neural-symbolic concept learner. It uses a symbolic reasoning process to bridge the learning of visual concepts and the semantic parsing of textual questions without any intermediate annotations except for the final answers.  We refer readers to \cite{mao2019neuro,yi2018neural} for more details on this model.

\noindent{\textbf{Scene Parsing}.} A scene parsing module develops an object-based representation for each image. Concretely, we adopt a pre-trained Mask R-CNN~\cite{he2017mask} to generate object proposals from the image. The detected bounding boxes with the original image are sent to a ResNet-34~\cite{he2016deep} to extract the object-based features. 

\noindent{\textbf{Concept Embeddings}.} By assuming each visual attribute (\eg, shape) contains a set of visual concepts (\eg, cylinder), the extracted visual features are embedded into concept spaces by learnable neural operators of the attributes. 

\noindent{\textbf{Question Parsing}.} The question parsing module translates a question in natural language into an executable program in a domain-specific language designed for VQA. The question parser generates the latent program from a question in a sequence-to-sequence manner. A bi-directional LSTM is used to encode the input question into a fixed-length representation. The decoder is an attention-based LSTM, which produces the operations in the program step-by-step. Some operations take concepts as their parameters, such as \textit{Filter[Cube]} and \textit{Relate[Left]}. These concepts are selected from the concepts appearing in the question by the attention mechanism.

\noindent{\textbf{Symbolic Reasoning}.} Given the latent program, the symbolic executor runs the operations in the program with the object-based image representation to derive an answer for the input question. The execution is fully differentiable with respect to the concept embeddings since the intermediate results are represented in a probabilistic manner. Specifically, we keep an attention mask on all object proposals, with each element in the mask denoting the probability that the corresponding object contains certain concepts. The attention mask is fed into the next operation, and the execution continues. The final operation predicts an answer to the question. We refer the readers to the supplementary materials for more details and examples of the symbolic execution process.

\noindent{\textbf{Joint Optimizing}.} We formulate the problem of jointly learning the question parser and the concept embeddings without the annotated programs. Suppose we have a training sample consisting of image $I$, question $Q$, and answer $A$, and we do not observe the latent program $l$. The goal of training the whole system is to maximize the following conditional probability:
\begin{align}
    p(A|I, Q) = \E_{l \sim p(l|Q)}~[p(A|l,I)], \label{eq:prob}
\end{align}
where $p(l|Q)$ is parametrized by the question parser with the parameters $\theta_l$ and $p(A|l,I)$ is parametrized by the concept embeddings $\theta_e$ (there are no learnable parameters in the symbolic reasoning module). 
Considering the expectation over the program space in Eq.~\ref{eq:prob} is intractable, we approximate the expectation with Monte Carlo sampling. Specifically, we first sample a program $\hat{l}$ from the question parser $p(l|Q;\theta_l)$ and then apply $\hat{l}$ to obtain a probability distribution over possible answers $p(A|\hat{l}, I; \theta_e).$

Recalling the program execution is fully differentiable  w.r.t. the concept embeddings, we learn the concept embeddings by directly maximizing $\log p(A|\hat{l}, I; \theta_e)$ using gradient descent and the gradient $\nabla_{\theta_e}\log p(A|\hat{l}, I; \theta_e)$ can be calculated through back-propagation. Since the hard selection of $\hat{l}$ through Monte Carlo sampling is non-differentiable, the gradients of the question parser cannot be computed by back-propagation. Instead we optimize the question parser using the REINFORCE algorithm~\cite{williams1992simple}. The gradient of the reward function $J$ over the parameters of the policy is:
\begin{align}
\nabla J({\theta_l})=\mathbb{E}_{{l \sim p(l|Q; \theta_l)}}\left[\nabla \log p\left(l|Q ; \theta_l\right) \cdot r\right],
\end{align}
where $r$ denotes the reward. Defining the reward as the log-probability of the correct answer and again, we rewrite the intractable expectation with one Monte Carlo sample $\hat{l}$: 
\begin{align}
\nabla J({\theta_l})=\nabla \log p\left(\hat{l}|Q ; \theta_l\right) \cdot [\log p(A|\hat{l}, I; \theta_e) - b],
\end{align}
where $b$ is the exponential moving average of $\log p(A|\hat{l}, I; \theta_e)$, serving as a simple baseline to reduce the variance of gradients. Therefore, the update to the question parser at each learning step is simply the gradient of the log-probability of choosing the program, multiplied by the probability of the correct answer using that program.

\subsection{Background of Item Response Theory (IRT)}
Item response theory (IRT)~\cite{baker2001basics,baker2004item} was initially created in the fields of educational measurement and psychometrics. It has been widely used to measure the latent abilities of subjects (\eg, human beings, robots or AI models) based on their responses to items (\eg, test questions) with different levels of difficulty. The core idea of IRT is that the probability of a correct response to an item can be modeled by a mathematical function of both individual ability and item characteristics. 
 More formally, if we let $i$ be an individual and $j$ be an item, then the probability that the individual $i$ answers the item $j$ correctly can be modeled by a logistic model as:
\begin{align} \label{Eq:3PL}
    p_{ij} = c_j + \frac{1 - c_j}{1 + e^{-a_j(\theta_i - b_j)}},
\end{align}
where $\theta_i$ is the latent ability of the individual $i$ and $a_j, b_j, c_j$ are the characteristics of the item $j$. The item parameters can be interpreted as changing the shape of the standard logistic function: $a_j$ (the discrimination parameter) controls the slope of the curve; $b_j$ (the difficulty parameter) is the ability level, it is the point on $\theta_i$ where the probability of a correct response is the average of $c_j$ (min) and 1 (max), also where the slope is maximized; $c_j$ (the guessing parameter) is the asymptotic minimum of this function, which accounts for the effects of guessing on the probability of a correct response for a multi-choice item. Equation~\ref{Eq:3PL} is often referred to as the three-parameter logistic (3PL) model since it has three parameters describing the characteristics of items. We refer the readers to \cite{baker2001basics,baker2004item,embretson2013item} for more background and details on IRT.

\subsection{Multi-dimensional IRT using Model Responses}
Traditional IRT is proposed to model the human responses to several hundred items. However, datasets used in machine learning, especially deep neural networks, often consist of hundreds of thousands of samples or even more. It is costly to collect human responses for large datasets, and more importantly, human responses are not distinguishable enough to estimate the sample difficulties since samples in machine learning datasets are usually straightforward for humans. Lalor \etal~\cite{Lalor2016BuildingAE,lalor2019learning} empirically shows on two NLP tasks that IRT models can be fit using machine responses by comparing item parameters learned from the human responses and the responses from an artificial crowd of thousands of machine learning models.

Similarly, we propose to fit IRT models with accumulated model responses (\ie, the predictions of model snapshots) from the training process. Considering the compositional nature of visual reasoning, we propose a multi-dimensional IRT (mIRT) model to estimate the concept difficulty and model competence (corresponding to the subject ability in original IRT), from which the question difficulty can be further calculated.

Formally, we have $C$ concepts, $M$ model snapshots saved from all time steps, and $N$ questions. Let $\Theta = \{\theta_{ic}\}_{i=1..M}^{c=1...C}$, where $\theta_{ic}$ is the $i$-th snapshot's competence on the $c$-th concept, and $B = \{b_c\}^{c=1...C}$, where $b_c$ is the difficulty of the $c$-th concept, $\mathcal{Q}=\{q_{jc}\}^{c=1...C}_{j=1...N}$, where $q_{jc}$ is the number of the $c$-th concept in the $j$-th question and $g_j$ is the probability of guessing the correct answer to the $j$-th question, $\mathcal{Z} = \{z_{ij}\}_{i=1...M}^{j=1...N}$, where $z_{ij} \in \{0,1\}$ be the response of the $i$-th snapshot to the $j$-th question ($1$ if the model answers the question correctly and $0$ otherwise). The probability that the snapshot $i$ can correctly recognize the concept $c$ is formulated by a logistic function:

\begin{align} \label{Eq:concept_prob}
p_{ic}(\theta_{ic}, b_c) = \frac{1}{1+e^{-(\theta_{ic} - b_c)}}.
\end{align}
Then the probability that the snapshot $i$ answers the question $j$ correctly is calculated as:
\begin{align} \label{Eq:mIRT}
&p(z_{ij} = 1|\theta_{i}, B) = g_j + (1-g_j)\prod_{c=1}^C p_{ic}^{q_{jc}} .
\end{align}
The probability that the snapshot $i$ answers the question $j$ incorrectly is:
\begin{align}
&p(z_{ij}=0|\theta_{i}, B) = 1 - p(z_{ij}=1|\theta_{i}, B).
\end{align}
 The total data likelihood is:

\begin{align}
&p(\mathcal{Z}|\Theta, B) = \prod_{i=1}^{M}\prod_{j=1}^{N} p(z_{ij}|\theta_{i}, B).
\end{align}
This formulation is also referred to as conjunctive multi-dimensional IRT~\cite{reckase1985difficulty,reckase2009multidimensional}.

\subsection{Variational Bayesian Inference for mIRT}
The goal of fitting an IRT model on observed responses is to estimate the latent subject abilities and item parameters. In traditional IRT, the item parameters are usually estimated by Marginal Maximum Likelihood (MML) via an Expectation-Maximization (EM) algorithm~\cite{bock1981marginal}, where the subject ability parameters are randomly sampled from a normal distribution and marginalized out. Once the item parameters are estimated, the subject abilities are scored by maximum a posterior (MAP) estimation based on their responses to items. However, the EM algorithm is not computational efficient on large datasets. One feasible way for scaling up is to perform variational Bayesian inference on IRT~\cite{natesan2016bayesian,lalor2019learning}. The posterior probability of the parameters in mIRT can be written as:
\begin{align} \label{Eq:posterior}
p(\Theta, B|\mathcal{Z}) = \frac{p(\mathcal{Z}|\Theta, B)p(\Theta) p(B)}{\int_{\Theta, B} p(\Theta, B, \mathcal{Z})},
\end{align}
where $p(\Theta),p(B)$ are the priors distribution of $\Theta$ and $B$. The integral over the parameter space in Eq~\ref{Eq:posterior} is intractable. Therefore, we approximate it by a factorized variational distribution on top of an independence assumption of $\Theta$ and $B$:
\begin{align}
q(\Theta, B)=\prod_{i=1, c=1}^{M, C} \pi_{ic}^{\theta}\left(\theta_{ic}\right) \prod_{c=1}^{C} \pi_{c}^{b}\left(b_{c}\right),
\end{align}
where $\pi_{ic}^\theta$ and $\pi_c^b$ denote Gaussian distributions for model competences and concept difficulties, respectively. We adopt the Kullback-Leibler divergence (KL-divergence) to measure the distance of $p$ from $q$, which is defined as:
\begin{align} \label{Eq:KL}
D_{\mathrm{KL}}(q\|p) := \E_{q(\Theta, B)} \log\frac{q(\Theta, B)}{p(\Theta, B|\mathcal{Z})},
\end{align}
where $p(\Theta, B|\mathcal{Z})$ is still intractable. We can further decompose the KL-divergence as:
\begin{align}
D_{\mathrm{KL}}(q\| p)=\E_{q(\Theta, B)} \left[\log\frac{q(\Theta, B)}{p(\Theta, B,\mathcal{Z})} + \log p(\mathcal{Z}) \right].
\end{align}
In other words, we also have:
\begin{align}
\log p(\mathcal{Z}) &= D_{\mathrm{KL}}(q \| p)-\E_{q(\Theta, B)} \log\frac{q(\Theta, B)}{p(\Theta, B,\mathcal{Z})}\\
&=D_{\mathrm{KL}}(q \| p)+\mathcal{L}(q) .
\end{align}
As the log evidence $\log p(\mathcal{Z})$ is fixed with respect to $q$, maximizing the final term $\mathcal{L}(q)$ minimizes the KL divergence of $q$ from $p$. And since $q(\Theta, B)$ is a parametric distribution we can sample from, we can use Monte Carlo sampling to estimate this quantity. Since the KL-divergence is non-negative, $\mathcal{L}(q)$ is an evidence lower bound (ELBO) of $\log p(\mathcal{Z})$. By maximizing the ELBO with an Adam optimizer~\cite{kingma2014adam} in Pyro~\cite{bingham2018pyro}, we can estimate the parameters in mIRT.

\subsection{Training Samples Selection Strategy}
The proposed model can estimate the question difficulty for the current model competence without looking at the ground-truth images and answers. It facilitates the active selection for future training samples. More specifically, we can easily calculate the probability that the model answers a given question correctly from Eq.~\ref{Eq:concept_prob} and Eq.~\ref{Eq:mIRT} (without guessing) using estimated $\Theta$ and $b$. This probability serves as an indicator of the question difficulty for the learner in each stage. The higher the probability, the easier the question. To select appropriate training samples, we rank the questions and filter out the hardest questions by setting a probability lower bound (LB) and the easiest questions by a probability upper bound (UB). Algorithm~\ref{alg:training} summarizes the overall training process. We will discuss the influence of LB and UB on the learning process in Section~\ref{sec:exp_selection}.

\begin{algorithm}[H]
    \caption{Competence-aware Curriculum Learning}
    \label{alg:training}
    \begin{algorithmic}[l]
    \STATE \textbf{Initialization}:
    the training set $\mathcal{D}=\{(I_j, Q_j, A_j)\}_{j=1}^N$,
    concept difficulty $B^{(0)}$, model competence $\Theta^{(0)}$,
    concept learner $\phi^{(0)}$, accumulated responses $\mathcal{Z} = \{\}$
    \FOR{$t = 1~\text{to}~T$}
        \STATE $\Theta^{(t)}, B^{(t)} = \arg\min_{\Theta, B} \mathcal{L}(q;\Theta^{(t-1)}, B^{(t-1)}, \mathcal{Z})$
        \STATE $\mathcal{D}^{(t)} = \{(I, Q, A) : \mathrm{LB} \leq p(Q; \Theta^{(t)}, B^{(t)}) \leq \mathrm{UB}\}$
        \STATE $\phi^{(t)}, \mathcal{Z}^{(t)} = \mathrm{Train}(\phi^{(t-1)}, \mathcal{D}^{(t)}$)
        \STATE $\mathcal{Z} = \mathcal{Z} \cup \mathcal{Z}^{(t)} $
    \ENDFOR
    \end{algorithmic}
\end{algorithm}
\section{Experiments} \label{sec:exp}

\subsection{Experimental Setup}
\noindent{\textbf{Dataset}}. We evaluate the proposed method on the CLEVR dataset~\cite{johnson2017clevr}, which consists of a training set of 70k images and $\sim$700k questions, and a validation set of 15k images and $\sim$150k questions. The proposed model selects questions from the training set during learning, and we evaluate our model on the entire validation set. 

\noindent{\textbf{Models}}. To analyze the performance of the proposed approach, We conduct experiments by comparing with several model variants:
\begin{itemize}[leftmargin=*,noitemsep,nolistsep]
    \item \textbf{FiLM-LBA}: the best model from~\cite{Misra2017LearningBA}.
    \item \textbf{NSCL}: the neural-symbolic concept learner~\cite{mao2019neuro} without using any curriculum. Questions are randomly sampled from the training set.
    \item \textbf{NSCL-Fixed}: NSCL following a manually-designed discretized curriculum.
    \item \textbf{NSCL-mIRT}: NSCL following a continuous curriculum built by the proposed mIRT estimator.
\end{itemize}

Please refer to the supplementary materials for detailed model settings and learning techniques during training.

\subsection{Training Process \& Model Performance}

\begin{figure}[t]
    \centering {\includegraphics[width=1\textwidth]{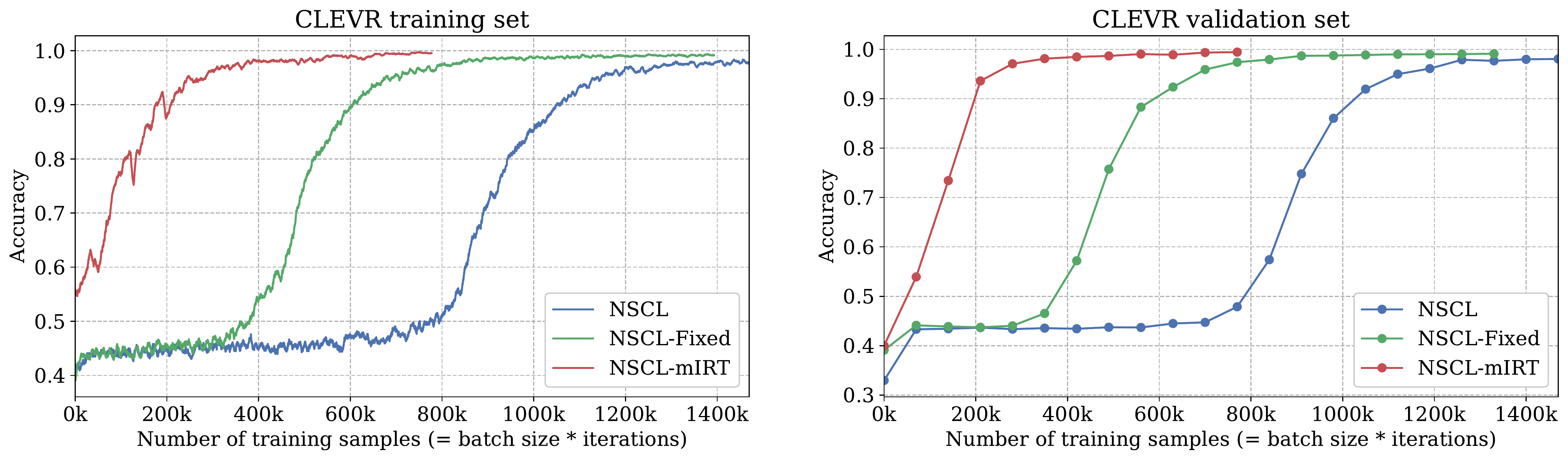}}
    \caption{The learning curves of different model variants on the CLEVR dataset.}
    \label{fig:vqa_acc}
\end{figure}

Figure~\ref{fig:vqa_acc} shows the accuracies of the model variants at different timesteps on the training set (left) and validation set (right). Notably, the proposed NSCL-mIRT converges almost 2 times faster than NSCL-Fixed and 3 times faster than NSCL (\ie, 400k v.s. 800k v.s. 1200k). Although NSCL-mIRT spends extra time to estimate the parameters of the mIRT model, such time cost is negligible compared to other time spent in training (less than 1\%). From \autoref{tab:acc}, we can see that NSCL-mIRT consistently outperforms FilM-LBA at various iterations, which demonstrates the preeminence of mIRT in building an adaptive curriculum.

Besides, NSCL-mIRT consumes less than 300k unique questions for training when it converges. It indicates that NSCL-mIRT saves about 60\% of the training data, which largely eases the data redundancy problems. It provides a promising direction for designing a data-efficient curriculum and helping current data-hungry deep learning models save time and money cost during data annotation and model training.

Moreover, NSCL-mIRT obtains even higher accuracy than NSCL and NSCL-Fixed. This indicates that the adaptive curriculum built by the multi-dimensional IRT model not only remarkably increases the speed of convergence and reduces the data consumption during the training process, but also leads to better performance, which also verifies the hypothesis made by Bengio \etal~\cite{bengio2009curriculum}.

\begin{figure}[t]
    \centering {\includegraphics[width=1\textwidth,height=0.3\textwidth]{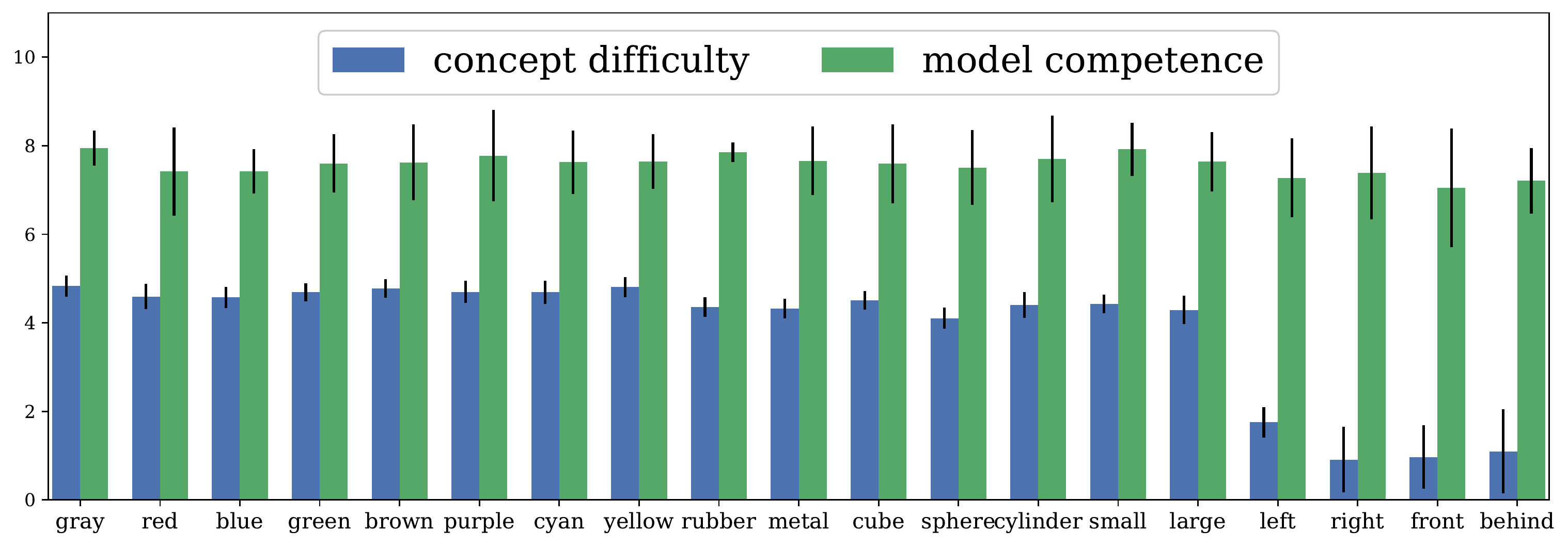}}
    \caption{The estimated concept difficulty and model competence at the final iteration.}
    \label{fig:irt_estimation}
\end{figure}

\begin{figure}[t]
    \centering \subfigure[]{\includegraphics[width=0.49\textwidth]{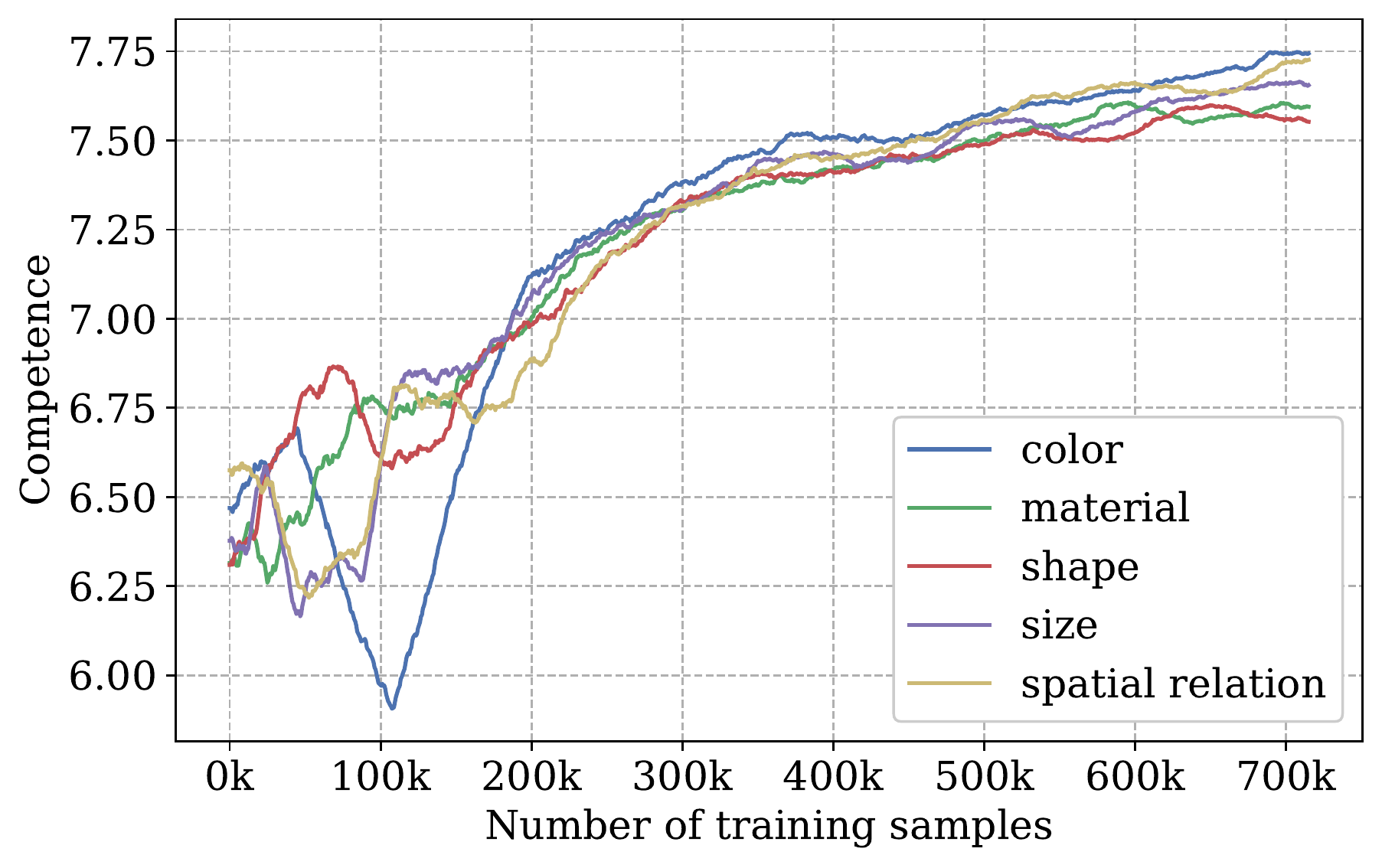}\label{fig:competence}} 
    \subfigure[]{\includegraphics[width=0.49\textwidth]{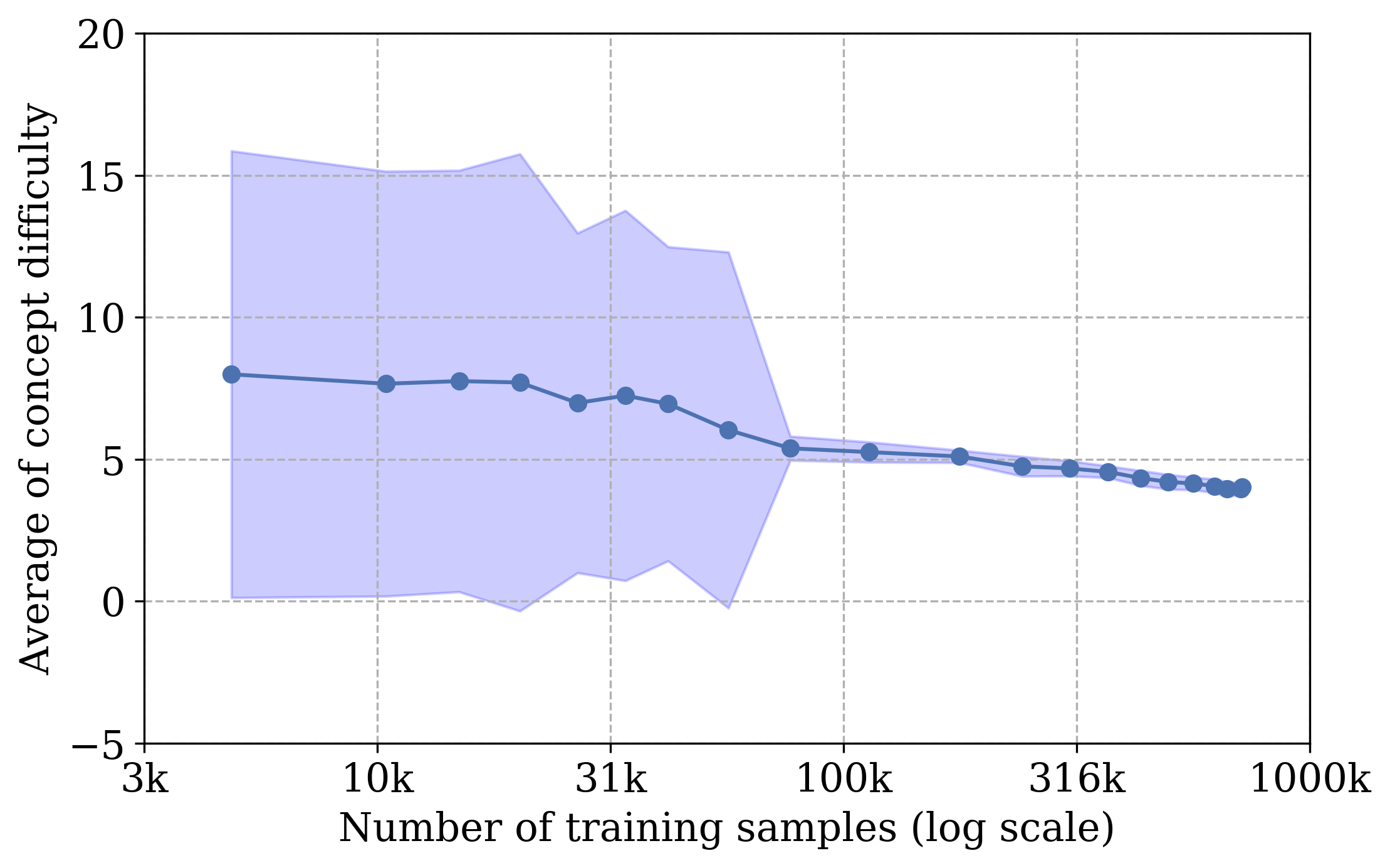}\label{fig:loc_scale_diff}}
    \caption{(a) The estimated model competence at various iterations for different attributes. The value for each attribute type is averaged from the visual concept it contains.
    (b) The estimated concept difficulty at various iterations. The shaded area represents the variance of the estimations.}
\end{figure}

\subsection{Multi-dimensional IRT}

The estimated concept difficulty and model competence after converging is shown in Figure~\ref{fig:irt_estimation} for studying the performance of the mIRT model. Several critical observations are: 
(1) The spatial relations (\ie, left/right/front/behind) are the easiest concepts. It satisfies our intuition since the model only needs to exploit the object positions to determine their spatial relations without dealing with appearance. The spatial relations are learned during the late stages since they appear more frequently in complex questions to connect multiple concepts.
(2) Colors are the most difficult concepts. The model needs to capture the subtle differences in the appearance of objects to distinguish eight different colors.
(3) The model competence scores surpass the concept difficulty scores for all the concepts. This result corresponds to the nearly perfect accuracy ($> 99\%$) on all questions and concepts.

Figure~\ref{fig:competence} shows the estimation of the model competence for each attribute type at various iterations. We can observe that model competence consistently increases throughout the training. Figure~\ref{fig:loc_scale_diff} shows the estimations of the concept difficulty at different learning steps. As the training progresses, the estimations become more stable with smaller variance since more model responses are accumulated.

\subsection{Concept Learner}

\begin{table}[t]
\centering
\begin{minipage}{.49\linewidth}
\caption{The VQA accuracy of different models on the CLEVR validation set at various iterations. NSCL and NSCL-Fixed continue to improve with longer training steps, which is not shown for space limit.}\label{tab:acc}
\resizebox{\columnwidth}{!}{

\begin{tabular}{l|llllllll}
\hline
{Models}     & 70k    &140k    &280k    &420k    &630k    &700k \\ \hline
{FiLM-LBA~\cite{Misra2017LearningBA}}   & 51.2 &\textbf{76.2}   &92.9   &94.8   &95.2   &97.3\\
{NSCL}       & 43.3 &43.4    &43.3    &43.4    &44.5    &44.7 \\ 
{NSCL-Fixed} & 44.1    &43.9    &44.0    &57.2    &92.4    &95.9 \\
{NSCL-mIRT}  & \textbf{53.9}    &73.4    &\textbf{97.1}    &\textbf{98.5}    &\textbf{98.9}    &\textbf{99.3} \\
\hline
\end{tabular}}
\end{minipage}~~~
\begin{minipage}{.49\linewidth}
\caption{The accuracy of the visual attributes of different models. Please refer to the supplementary materials for detailed performance on each visual concept (\ie, ``gray'' and ``red'' in color attribute).}\label{tab:concept_acc}
\resizebox{\columnwidth}{!}{
 \begin{tabular}{ l|c c c c c c} 
 \hline 
 Model & Overall & Color & Material & Shape & Size\\
 \hline
 IEP~\cite{johnson2017clevr} & 90.6 & 91.0 & 90.0 & 89.9 & 90.6 \\
 MAC~\cite{hudson2018compositional} & 95.9 & 98.0 & 91.4 & 94.4 & 94.2 \\
 NSCL-Fixed~\cite{mao2019neuro} & 98.7 & 99.0 & 98.7 & 98.1 & 99.1\\
 \hline
 NSCL-mIRT & \textbf{99.5} & \textbf{99.5} & \textbf{99.7} & \textbf{99.4} & \textbf{99.6}\\
 \hline
\end{tabular}}
\end{minipage}
\end{table}

\begin{table}[!t]
\centering
\begin{minipage}{.5\linewidth}
\caption{Comparisons of the VQA accuracy on the CLEVR validation set with other models. }\label{tab:vqa_acc}
\resizebox{\columnwidth}{!}{
 \begin{tabular}{ l|c c c c c c} 
 \hline 
 Model & Overall & Count & \begin{tabular}{@{}c@{}}Cmp \\ Num.\end{tabular} & Exist 
 & \begin{tabular}{@{}c@{}}Query \\ Attr.\end{tabular} 
 & \begin{tabular}{@{}c@{}}Cmp \\ Attr.\end{tabular}\\
 \hline
 Human & 92.6 & 86.7 & 86.4 & 96.6 & 95.0 & 96.0  \\ 
 \hline
 IEP~\cite{johnson2017clevr} & 96.9 & 92.7 & 98.7 & 97.1 & 98.1 & 98.9 \\
 FiLM~\cite{Perez2017FiLMVR} & 97.6 & 94.5 & 93.8 & 99.2 & 99.2 & 99.0 \\ 
 MAC~\cite{hudson2018compositional} & 98.9 & 97.2 & 99.4 & 99.5 & 99.3 & 99.5 \\
 NSCL~\cite{mao2019neuro} & 98.9 & 98.2 & 99.0 & 98.8 & 99.3 & 99.1\\
 NS-VQA~\cite{yi2018neural} & \textbf{99.8} & \textbf{99.7} & \textbf{99.9} & \textbf{99.9} & \textbf{99.8} & \textbf{99.8}\\
 \hline
 NSCL-mIRT & 99.5 & 98.9 & 99.0 & 99.7 & 99.7 & 99.6\\
 \hline
\end{tabular}}
\end{minipage}~~~
\begin{minipage}{.5\linewidth}
\caption{The VQA accuracy on CLEVR validation set with different LBs and UBs in the question selection strategy. Both LB and UB are in log scale.}\label{tab:LB_UB}
\resizebox{\columnwidth}{!}{
\begin{tabular}{l|llllll}
\hline
{(LB,UB)}      & 70k            & 140k           & 210k          & 280k           & 560k           & 770k \\
\hline
{(-10, 0)} & 44.39          & 52.01          & 63.04         & 73.5           & 97.93          & 99.01 \\
{(-5, 0)}  & 53.75          & 69.55          & 82.44         & 95.31          & 98.92          & 99.27 \\
{(-3, 0)}  & 51.38          & 55.97          & 58.33         & 65.11          & 69.57          & 70.01 \\
{(-5, -0.5)} & 42.06          & 52.67          & 80.46         & 95.54          & 98.41          & 99.06 \\
{(-5, -0.75)} & \textbf{53.91} & \textbf{73.42} & \textbf{93.6} & \textbf{97.07} & 99.04          & \textbf{99.50} \\
{(-5, -1)} & 44.57          & 63.65          & 82.95         & 94.38          & \textbf{99.15} & 99.48 \\ 
\hline
\end{tabular}}
\end{minipage} 
\end{table}

We apply the count-based concept evaluation metric proposed in~\cite{mao2019neuro} to measure the performance of the concept learner, which evaluates the visual concepts on synthetic questions with a single concept such as ``How many \textit{red} objects are there?"
Table~\ref{tab:concept_acc} presents the results by comparing with several state-of-the-art methods, which includes methods based on neural module network with programs (IEP~\cite{johnson2017clevr}) and neural attentions without programs~(MAC~\cite{Hu2017LearningTR}). Our model achieves nearly perfect performance across visual concepts and outperforms all other approaches. This means the model can learn visual concepts better with an adaptive curriculum. Our model can also be applied to the VQA. Table~\ref{tab:vqa_acc} summarizes the VQA accuracy on the CLEVR validation split. Our approach achieves comparable performance with state-of-the-art methods.

\subsection{Question Selection strategy}\label{sec:exp_selection}
The question selection strategy is controlled by two hyper-parameters: the lower bound (LB) and upper bound (UB). We conduct experiments by learning with different LBs and UBs, and Table~\ref{tab:LB_UB} shows the VQA accuracy at various iterations. It reveals that the proper lower bound can effectively filter out too hard questions and accelerate the learning at the early stage of the training, as shown in the first three rows. Similarly, a proper upper bound helps to filter out too easy questions at the late stage of the training when the model has learned most concepts. 
Please refer to the supplementary material for the visualization of selected questions at various iterations.

\vspace{-0.1cm}
\section{Conclusions and Discussions} \label{sec_discussion}
\vspace{-0.1cm}

We propose a competence-aware curriculum for visual concepts learning via question answering. We design a multi-dimensional IRT model to estimate concept difficulty and model competence at each training step from the accumulated model responses generated by different model snapshots. The estimated concept difficulty and model competence are further used to build an adaptive curriculum for the visual concept learner. Experiments on the CLEVR dataset show that the concept learner with the proposed competence-aware curriculum converges three times faster and consumes only $40\%$ of the training data while achieving similar or even higher accuracy compared with other state-of-the-art models. 

In the future, our work can be potentially applied to \textit{real-world images} like GQA~\cite{hudson2019gqa} and VQA-v2~\cite{goyal2017making} datasets, by explicitly modeling the relationship among visual concepts. However, there are still unsolved challenges for real-world images. Specifically, compared with synthetic images in CLEVR, real-world images have a much larger vocabulary of visual concepts. For example, as shown in \cite{anderson2018bottom}, there are over 2,000 visual concepts in MSCOCO images. Usually, these concepts are automatically mined from image captions and scene graphs. Thus some of them are highly correlated like ``huge'' and ``large'', and some of them are very subjective like ``busy'' and ``calm''. Such a large and noisy vocabulary of visual concepts is challenging for the mIRT model since current visual concepts are assumed to be independent. It also requires a much longer time to converge when maximizing the ELBO to fit the mIRT model with more concepts. A potential solution is to consider the hierarchical structure of visual concept space and correlations among the concepts and incorporate commonsense knowledge to handle subjective concepts.

More importantly, the competence-aware curriculum can be adapted to other domains that possess compositional structures such as natural language processing. Specifically, in neural machine translation task~\cite{sutskever2014sequence,bahdanau2015neural}, mIRT can be used to model the difficulty and competence of translating different words/phrases and build a curriculum to increase learning speed and data efficiency. mIRT can also be used in the task of semantic parsing~\cite{dong2016language,Liang2016-qo,Liang2018-aq} that transforms natural language sentences (\eg, instructions or queries) into logic forms (\eg, lambda-calculus or SQL). The difficulty and competence of different logic predicates can also be estimated by the mIRT model.

\paragraph{\textbf{Acknowledgements.}}
We thank Yixin Chen from UCLA for helpful discussions. This work reported herein is supported by ARO W911NF1810296, DARPA XAI N66001-17-2-4029, and ONR MURI N00014-16-1-2007.

%
%
\bibliographystyle{splncs04}
\bibliography{main}

\end{document}


\pagestyle{headings}
\mainmatter
\def\ECCVSubNumber{3029}  

\title{Supplementary Materials for \\ A Competence-aware Curriculum for Visual Concepts Learning via Question Answering} 

\titlerunning{Competence-aware Curriculum for Visual Concepts Learning}
\author{Qing Li\orcidID{0000-0003-1185-5365} \and
Siyuan Huang\orcidID{0000-0003-1524-7148} \and
Yining Hong\orcidID{0000-0002-0518-2099} \and
Song-Chun Zhu\orcidID{0000-0002-1925-5973}}
\authorrunning{Q. Li et al.}
%
\institute{UCLA Center for Vision, Cognition, Learning, and Autonomy (VCLA)
\\ \email{\{liqing, huangsiyuan, yininghong\}@ucla.edu}, \email{sczhu@stat.ucla.edu}
}

\maketitle

\section{Training Details}
\noindent{\textbf{Scene Parsing}:} Following~\cite{yi2018neural}, we generate object proposals with pre-trained Mask R-CNN. The Mask R-CNN module is pre-trained on 4k generated CLEVR images with bounding box annotations only. We use ResNet-50 FPN as the backbone and train the model for 30k iterations with a batch size of 8. 

\noindent{\textbf{Question Parser and Concept Embedding}:} For the question parser, both the encoder and decoder are LSTMs of two hidden layers with a 256-dim hidden vector. The dimension of the word embedding is 300. The question parser is pre-trained with 1k randomly selected question-program pairs. Please refer to Appendix A in~\cite{mao2019neuro} for the specification of the domain-specific language (DSL) designed for the CLEVR dataset to represent the programs.
Following~\cite{mao2019neuro}, we set the dimension of the concept embedding as 64. During the joint optimization of concept embedding and question parser, we adopt the Adam optimizer~\cite{kingma2014adam} with a fixed learning rate of 0.001, and the batch size is 64. 

\noindent{\textbf{Multi-dimensional IRT (mIRT)}:} The mIRT model is implemented in Pyro~\cite{bingham2018pyro}, which is a probabilistic programming framework using PyTorch as the backend. We train the mIRT model using an Adam optimizer with a learning rate of 0.1. The training of the mIRT model converges fast and usually in less than 1000 iterations, therefore the running time is negligible compared to the time of training the visual concept learner. 

\begin{figure}[t]
	\centering {\includegraphics[width=0.5\textwidth]{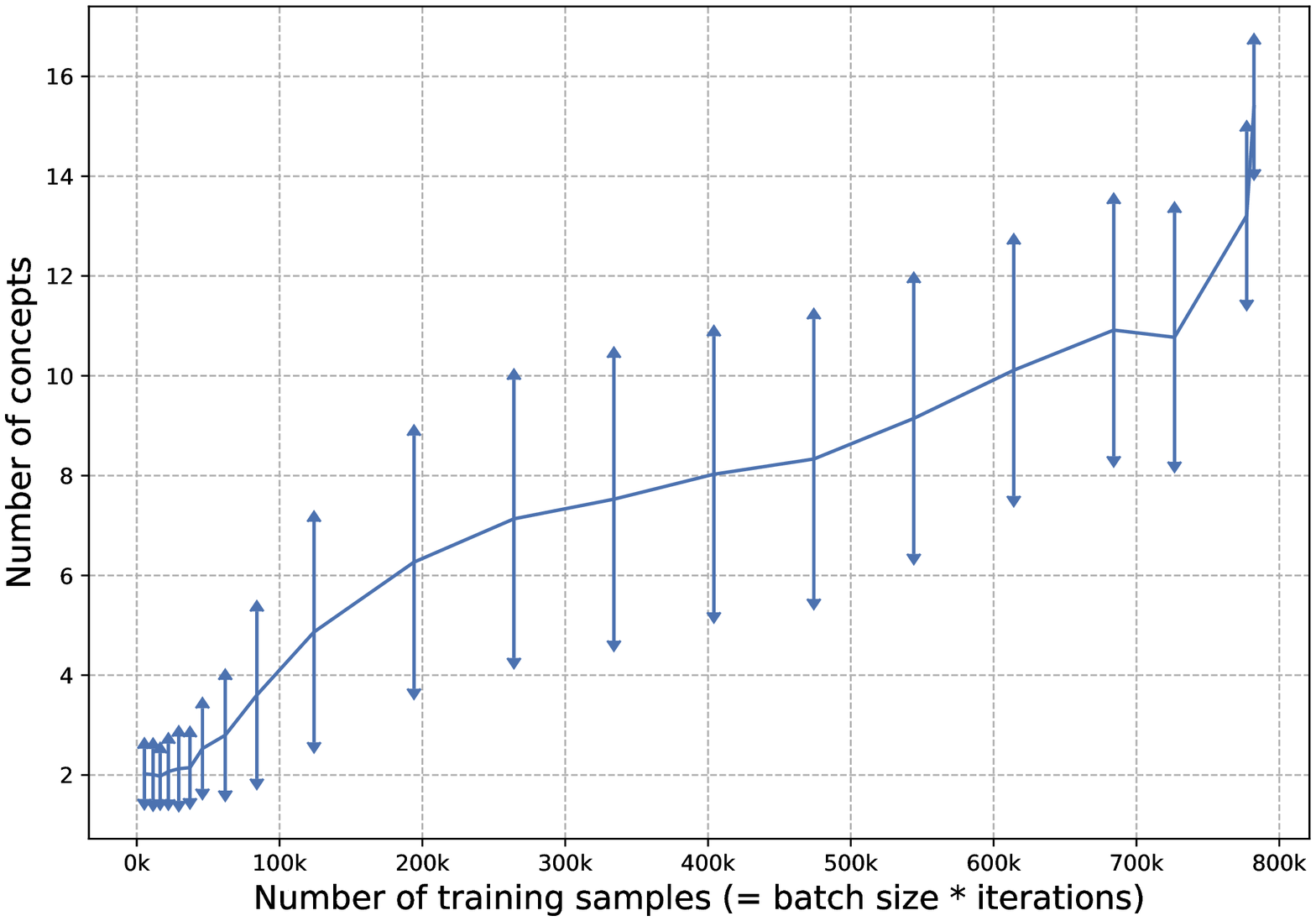}}
	\caption{The average number of concepts of selected questions smoothly increases during training, which suggests that the training follows an easy-to-hard curriculum.}
	\label{fig:n_concepts}
\end{figure}

\begin{figure*}[ht]
	\centering {\includegraphics[width=1\textwidth]{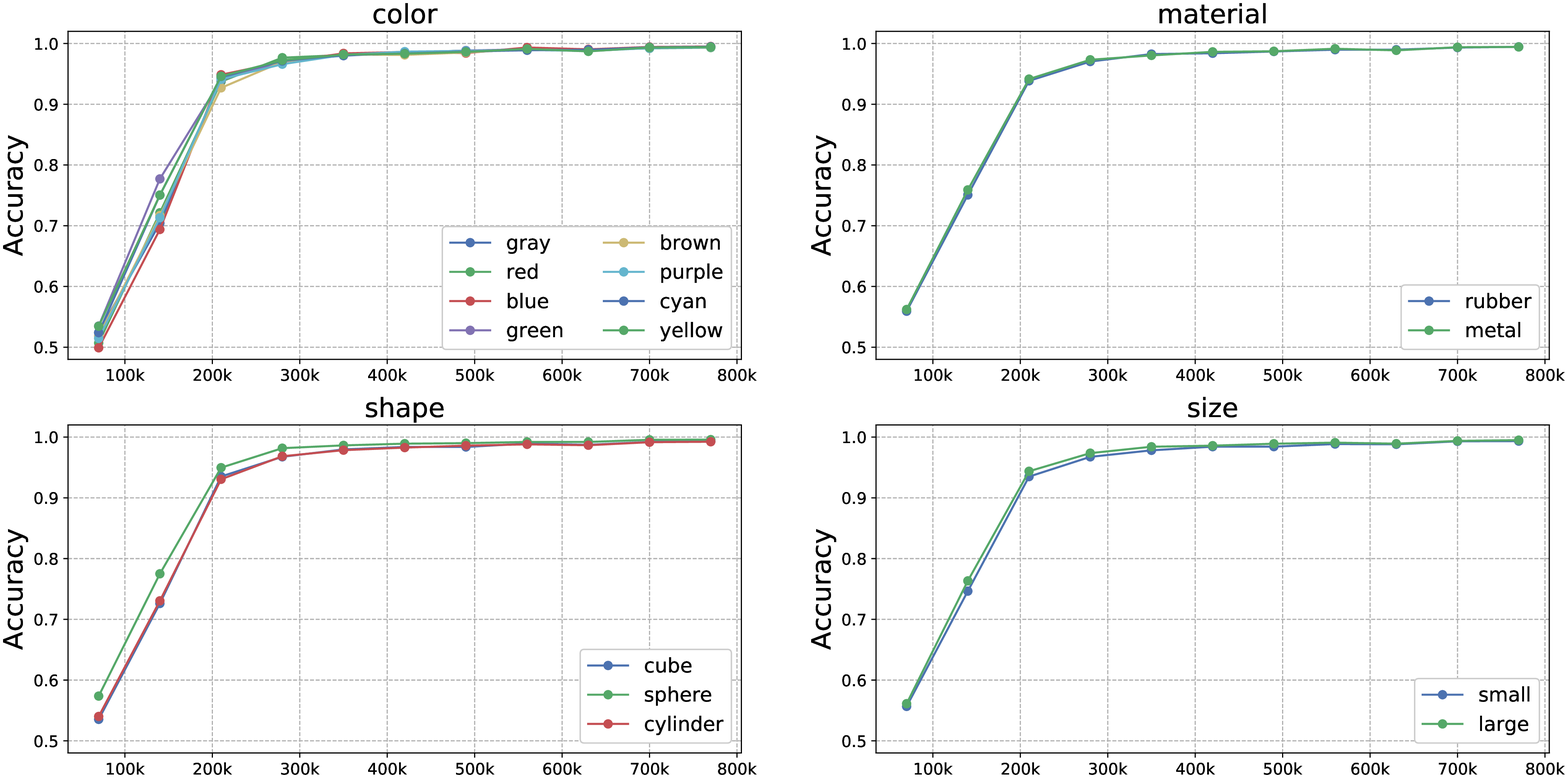}}
	\caption{The accuracy of each concept at various iterations. The concepts are grouped by the attribute type.}
	\label{fig:concept_acc}
\end{figure*}

\noindent{\textbf{Training Steps}:} The length of each training epoch is determined by the number of selected questions at this epoch. Questions are selected by the proposed training sample selection strategy, as illustrated in Section {\color{red}3.5}. We train the model from the easiest samples. Specifically, we select 5k samples with less than two concepts as the starting questions. As shown in Figure~\ref{fig:n_concepts}, the number of selected questions grows along with the increasing model competence. In the end, the model selected the few hardest questions and then converges, which also causes early stop since no question is selected in the next epoch. Similarly, Figure~\ref{fig:concept_acc} shows the accuracy of each concept at various iterations.

\noindent{\textbf{Training Speed}:} We train the model on a single Nvidia TITAN RTX card, and the entire convergence time is about 10 hours, with 21 epochs (about 11k iterations). All our models are implemented in PyTorch.



\section{Visualization of Selected Questions}

\begin{figure*}[t]
	\centering {\includegraphics[width=1\textwidth]{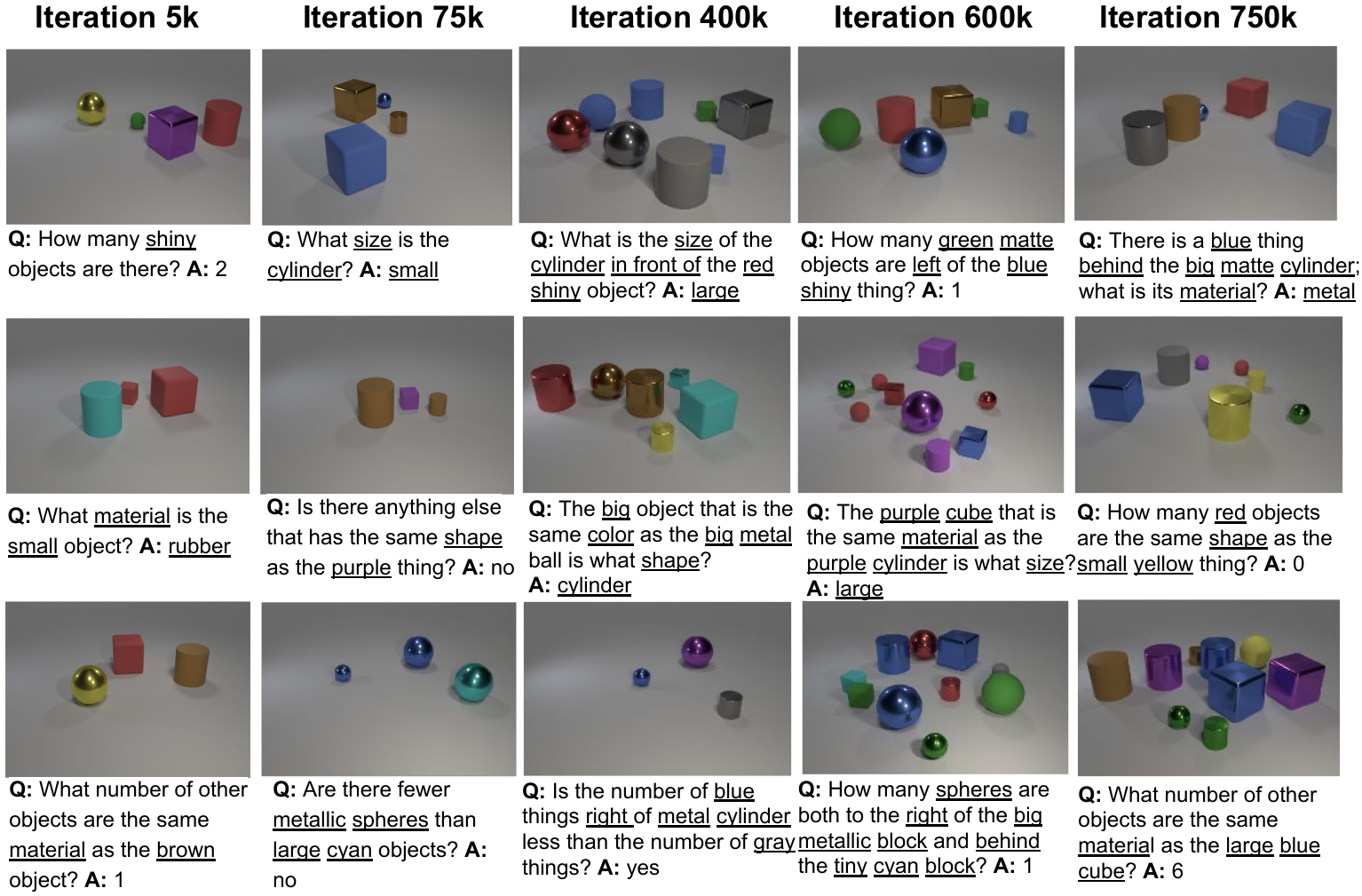}}
	\caption{Example questions selected at different iterations (LB=-5, UB=-0.75). The proposed model selects increasingly complex questions during the training progress. It starts the learning with simple questions with one or two concepts and moves to complex ones involving combined concepts with spatial relationships.}
	\label{fig:selected_examples}
\end{figure*}

Figure~\ref{fig:selected_examples} shows model responses for the selected questions at various iterations. They represent the smooth improvements for the question difficulty and model competence during the training process. Specifically, in the early stages of training, the model selects easy questions in simple scenes, which only involves one or two concepts. Following the increase of model competence, the selection strategy starts to tackle hard questions in complex scenes, consisting of multiple concepts with spatial relationships. Without any extra prior knowledge, this easy-to-hard learning process shows its smoothness and efficiency with automatic guidance from the proposed curriculum. 

\section{Qualitative Examples of NSCL}
\begin{figure}[ht]
	\centering {\includegraphics[width=1\textwidth]{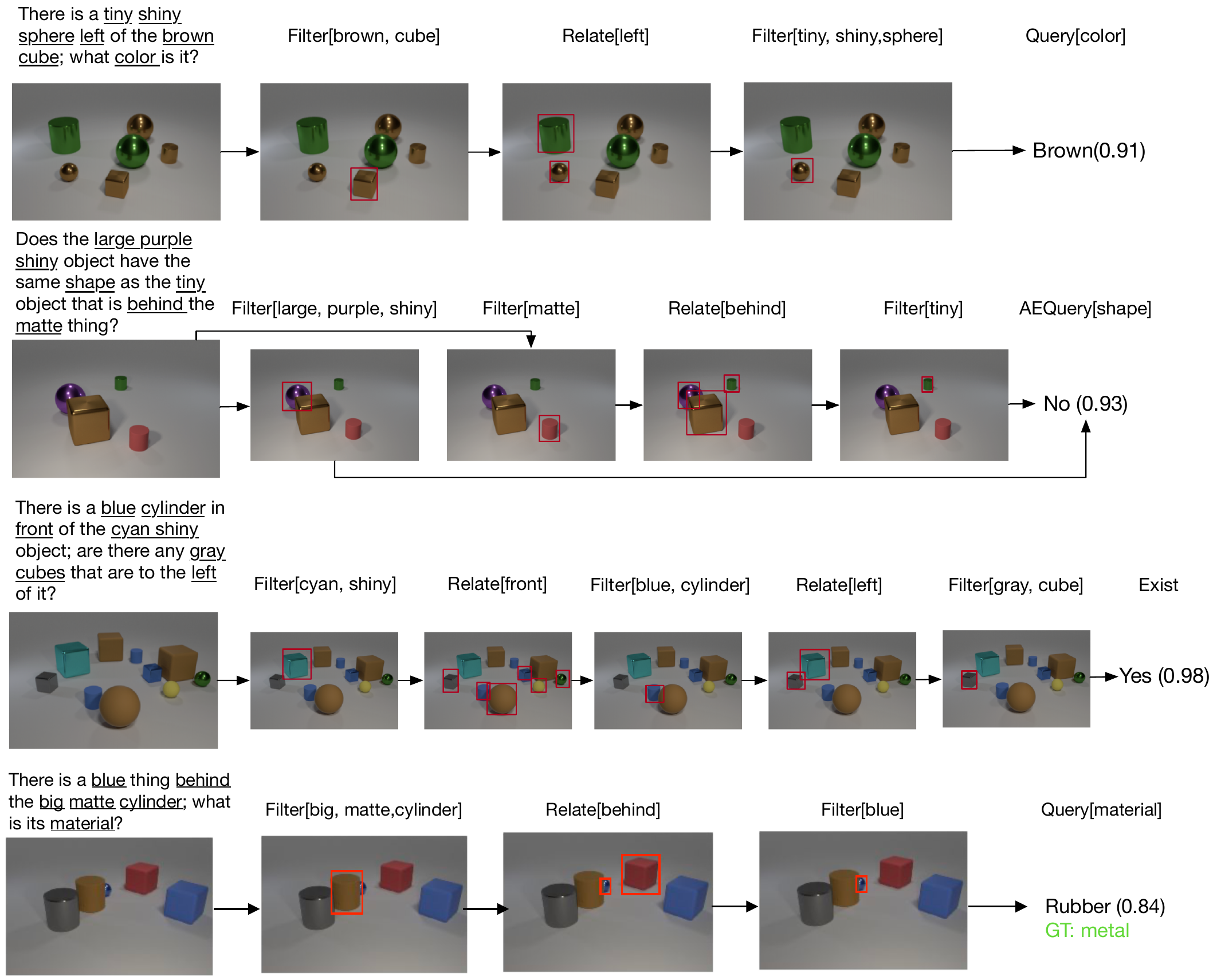}}
	\caption{Visualization of the symbolic reasoning process by the neural-symbolic concept learner on the CLEVR dataset. The questions of the first three examples are correctly answered by our model, and the last example is a typical error case caused by a small object under heavy occlusion.}
	\label{fig:symbolic_reasoning}
\end{figure}

Figure~\ref{fig:symbolic_reasoning} visualizes several examples of the symbolic reasoning process by the neural-symbolic concept learner. The questions of the first three examples are correctly answered by our model, and the last example is a typical error case caused by a small object under heavy occlusion.

\clearpage

%
%
\bibliographystyle{splncs04}
\bibliography{egbib}